\pdfoutput=1

\documentclass[11pt]{article}

\usepackage[final]{acl}
\usepackage{hhline}
\usepackage[linewidth=1pt]{mdframed}
\usepackage{tabularx}
\usepackage{array}
\usepackage{times}
\usepackage{latexsym}
\usepackage{booktabs}
\usepackage[T1]{fontenc}

\usepackage[utf8]{inputenc}

\usepackage{microtype}

\usepackage{inconsolata}

\usepackage{amssymb}
\usepackage{amsmath}
\usepackage{graphicx}
\usepackage{comment}
\usepackage{booktabs}
\usepackage{adjustbox}
\usepackage{multirow}
\usepackage{makecell}
\usepackage{algorithm}
\usepackage{algpseudocode}

\title{Can LLMs Understand the Implication of Emphasized Sentences \\in Dialogue?}

\author{Guan-Ting Lin, Hung-yi Lee\\
  Graduate Institute of Communication Engineering, National Taiwan University \\
  Taiwan \\
  \texttt{\{f10942104, hungyilee\}@ntu.edu.tw}\\}

\begin{document}
\maketitle
\begin{abstract}
Emphasis is a crucial component in human communication, which indicates the speaker's intention and implication beyond pure text in dialogue. While Large Language Models (LLMs) have revolutionized natural language processing, their ability to understand emphasis in dialogue remains unclear. This paper introduces \textbf{Emphasized-Talk}, a benchmark with emphasis-annotated dialogue samples capturing the implications of emphasis. We evaluate various LLMs, both open-source and commercial, to measure their performance in understanding emphasis. Additionally, we propose an automatic evaluation pipeline using GPT-4, which achieves a high correlation with human rating. Our findings reveal that although commercial LLMs generally perform better, there is still significant room for improvement in comprehending emphasized sentences\footnote{https://github.com/DanielLin94144/Emphasized-Talk}.

\end{abstract}

\section{Introduction}
Emphasis plays a key role in communication by highlighting important parts of dialogue. Recognizing these emphatic cues is crucial for achieving natural interactions, as they help reinforce intentions, and express nuances essential for understanding conversations. For instance, altering the emphasis in the sentence \textit{I never said he stole my bag} from \textit{he} to \textit{my} can drastically change its meaning. 
Emphasis can be represented in various ways, such as quotation marks, bold text, italics, capitalization, and underlining.

Large Language Models (LLMs)~\citep{llm, llama, gpt4} have revolutionized natural language processing by leveraging vast amounts of unlabeled text data. They can generate and understand natural language accurately, capturing nuances and complex relationships. As a result, LLMs are widely used in applications such as dialogue systems and virtual assistants.
Despite their capabilities, the ability of LLMs to grasp the \textit{subtle nuances and complex meanings conveyed through \textbf{emphasis} in human dialogue remains unclear}.  

\begin{figure}[t]{}
\centering\includegraphics[width=1\linewidth]{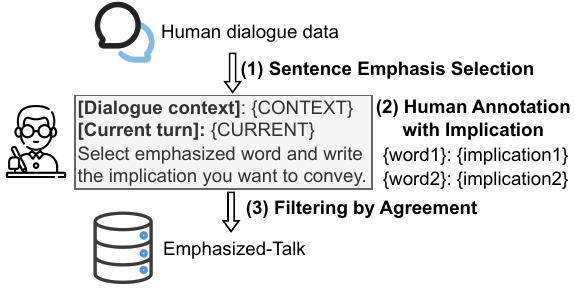}
    \caption{The illustration of Emphasized-Talk data collection pipeline. }
    \label{fig:data}
\end{figure}

To address this gap, we introduce a novel benchmark evaluation dataset, \textbf{Emphasized-Talk}, which features real dialogue samples with annotations capturing the implications of emphasis as interpreted by humans. Emphasized-Talk includes the same dialogue context and current sentence but with different words or phrases emphasized. This dataset is designed to test whether LLMs can accurately understand the meaning and intention behind emphasis in dialogue. 
Note that in this work, we use \texttt{quotation mark} ("") to highlight the emphasis in text sentences.
Our study includes both open-sourced and commercial LLMs of varying sizes to provide a comprehensive analysis of their current capabilities and limitations.

Furthermore, we propose an automatic evaluation method that leverages the capabilities of GPT-4~\cite{gpt4} to assess the performance of these models. By comparing various automatic evaluation methods, we find that GPT-4's evaluations correlate well with human evaluation scores. Our findings indicate that open-source LLMs generally struggle to understand the implications of emphasized sentences. While commercial LLMs perform better, there is still room for improvement in their comprehension of emphasized text.

Our contributions can be summarized as below: 
\begin{enumerate}
    \item We present a novel benchmark evaluation dataset, \textbf{Emphasized-Talk}, featuring real dialogues annotated with emphasis implications, providing a critical resource for testing LLMs' understanding of the emphasized text in conversations.
    \item Our study includes a comparative analysis of various LLMs, assessing their ability to interpret and generate emphasis in dialogue. This highlights the performance differences across models of different scales.
    \item We introduce an automated evaluation pipeline using GPT-4 with our benchmark, enabling efficient and consistent assessment of models, which significantly reduces the need for manual testing.
\end{enumerate}

\section{Related works}
\textbf{Human Communication Beyond Textual Information}:
In addition to text, the extra tags like emotion~\citep{pohl2017beyond, xue2023chat}, emoji \citep{demszky2020goemotions}, sentiment~\citep{10446933}, prosody~\citep{ward2012prosodic, lin2024advancing}, and non-lexical verbal sounds~\citep{ward2006non} are critical in understanding and generating human dialogue. These cues provide essential context and pragmatic use to the interaction, often conveying nuances that textual information alone cannot fully capture~\citep{ward2004pragmatic}. For instance, emotions and sentiment can indicate the speaker's attitude and feelings, while speaking style and prosody can reveal additional layers of meaning, such as sarcasm, emphasis, or urgency. This work focuses on emphasis as the non-textual cue in human communication, aiming to understand the implication meaning of beyond text. The quotation mark ("") is used to simulate the expression of emphasis. \\
\textbf{Emphasis in Dialogue}: 
In human communication, whether spoken or written, emphasis is a crucial tool for enhancing meaning, indicating emotional states, or highlighting important dialogue segments. In dialogue systems, understanding the subtleties conveyed through emphatic cues is essential for achieving truly natural interactions~\citep{pierrehumbert1990meaning, wagner2010experimental, jackson2016pragmatics}. Emphasis aids in disambiguating syntactic structures, reinforcing intentions, and expressing nuances that are vital for comprehensive understanding in conversational contexts~\citep{buchanan2013conversational}. Due to the importance of emphasis, studies on emphasis detection~\citep{arons1994pitch, suni2017hierarchical, talman-etal-2019-predicting, zhang2018emphasis, vaidya2022deep, morrison2024crowdsourced}, emphatic Text-to-Speech synthesis~\citep{fernandez2007automatic, seshadri22_interspeech, stephenson22_interspeech}, and Speech-to-Speech translation~\citep{goldman16_interspeech, emphassess} are long-standing research topics to capture emphasis. This work focus on evaluating the LLMs' ability to understand the high-level meaning and intention of emphasis.

\begin{table*}[t]
\centering
\begin{tabular}{lcccccccc}
\hline \hline
\multicolumn{1}{c|}{\multirow{2}{*}{\textbf{Model}}} & \multicolumn{2}{c|}{\textbf{MOS}}                     & \multicolumn{2}{c|}{\textbf{BERT$_{f1}$}}            & \multicolumn{2}{c|}{\textbf{auto-gpt4-gt}}            & \multicolumn{2}{c}{\textbf{auto-gpt4}}               \\ \cline{2-9} 
\multicolumn{1}{c|}{}                                & \multicolumn{1}{c}{score} & \multicolumn{1}{c|}{rank} & \multicolumn{1}{c}{score} & \multicolumn{1}{c|}{rank} & \multicolumn{1}{c}{score} & \multicolumn{1}{c|}{rank} & \multicolumn{1}{c}{score} & \multicolumn{1}{c}{rank} \\ \hline
ChatGPT                                              & 3.59                      &           2               & 88.7                      &        1                   & 2.94                      &           2                & 3.83                      &     2                      \\
Claude 3 Sonnet                                      & 3.73                      &          1                & 88.5                      &       2                    & 3.03                      &           1                & 3.86                      &     1                      \\
Llama 3-70B-chat                                     & 3.51                      &            3              & 88.2                      &       3                    & 2.79                      &           3                & 3.39                      &     3                      \\
Llama 2-70B-chat                                     & 3.21                      &           6               & 86.5                      &       6                    & 2.54                      &           4                & 2.92                      &     5                      \\
Llama 3-8B-instruct                                  & 3.41                      &           4               & 87.7                      &       4                   & 2.36                      &          6                 & 3.06                      &     4                      \\
Llama 2 7B-chat                                      & 2.61                      &           7               & 86.3                      &        7                   & 1.71                      &          7                 & 1.85                      &     7                      \\
Mistral 7B-instruct                                  & 3.29                      &         5                 & 87.4                      &       5                    & 2.47                      &          5                 & 2.82                      &     6                      \\ \midrule
Spearman's rank corr coef                         & \multicolumn{2}{c}{-}     & \multicolumn{2}{c}{0.964}                                  & \multicolumn{2}{c}{0.857}                                  & \multicolumn{2}{c}{0.964}                                  \\ 
     $p$-value                    & \multicolumn{2}{c}{-}     & \multicolumn{2}{c}{$4.5\times10^{-4}$}                    & \multicolumn{2}{c}{$1.4\times10^{-2}$}                    & \multicolumn{2}{c}{$4.5\times10^{-4}$}                                       
\\
\hline \hline   
\end{tabular}
\caption{The MOS, BERTscore, auto-gpt4-gt, and auto-gpt4 score of different LLMs. }
\label{tab:main}
\end{table*}
\section{Dataset: Emphasized-Talk}
The same context and current text with different emphasized words can lead to different implied meanings. Since no existing dataset contains varying emphasized words and meanings, we have created the first such dataset. Figure \ref{fig:data} illustrates the data collection. To build a real-world dialogue dataset with emphasized sentences, we initially used the DailyTalk~\citep{dailytalk} dataset, an open-domain multi-turn spoken dialogue dataset, as our content source. We then adopt the following strategies to create the Emphasized-Talk data:

\subsection{Sentence Emphasis Selection}

For each dialogue, we select the current sentence to be emphasized based on the following criteria:  \textbf{(1) Sufficient dialogue context}: The current turn is selected only if there are more than two preceding dialogue turns, ensuring adequate contextual information.   \textbf{(2) Availability of emphasized targets}: Since emphasizing function words like punctuation, articles, and proper nouns rarely affect sentence meaning, we select sentences containing more than four non-functional words and proper nouns for further consideration.

\subsection{Human Annotation with Implication}
Selecting \textit{which words and phrases to emphasize} is non-trivial. In human communication, emphasis placement depends on the dialogue context and the information the speaker wishes to convey. In this work, \textbf{\textit{human annotators choose where to place emphasis and document the implied meaning behind their choices}}, making the data more pragmatic and reflective of real-world dialogue. On average, the emphasized fragments consist of 1.15 words, indicating that annotators mostly emphasize single words, with occasional emphasis on phrases. Each annotator selects two different words or phrases to emphasize and note the implied meaning. The task template and instructions are shown in Appendix Figure \ref{fig:annotation}.

\begin{figure}[t]{}
\centering\includegraphics[width=0.8\linewidth]{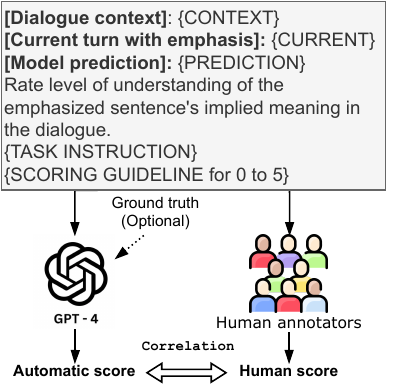}
\caption{Illustration of automatic and human evaluation of the model's predicted implications.}
\label{fig:eval}
\end{figure}

\subsection{Filtering by Agreement}
When multiple annotators select the same word or phrase to emphasize, there are often several annotations of implied meanings. If the implied meanings are significantly different between annotators, indicating a lack of consensus or too many possible implied meanings based on annotators' backgrounds, we check the semantic similarity among the annotated meanings using GPT-4 (see Appendix \ref{appendix:agreement_check} for details). Samples without agreement are filtered out. The percentage of samples that passed the filtering is 79.1\%. Overall, we collect 984 dialogue sample pairs. We show some examples in Appendix Table \ref{tab:filtering_agreement}.

\begin{table*}[t]
\centering
\adjustbox{width=0.9\textwidth}{
\begin{tabular}{ll}
\toprule
\textbf{Context}      & \begin{tabular}[c]{@{}l@{}}0: Umm excuse me, are there any more shopping carts?\\ 1: Yes, you can find it at the entrance.\end{tabular}                                                 \\ \hline
\textbf{Current turn} & 0: but there isn't a single one right now.                                                                                                                                  \\ \hline
\textbf{Predicted implication}   & \begin{tabular}[c]{@{}l@{}}``single": The customer needs an available shopping cart for individual use.\\ ``right now": There are no available shopping carts at the moment.\end{tabular} \\ \hline
\textbf{Ground truth implication} & \begin{tabular}[c]{@{}l@{}}``single": Not even one cart is available.\\ ``right now": The lack of carts is an immediate issue.\end{tabular}   \\ \bottomrule                                           
\end{tabular}}
\caption{Qualitative example with Emphasizing on different targets in the current turns. 0 and 1 denote the speaker's identity. For implication rows, the word or phrase within the quotation mark refers to the emphasized target, and the sentence after the colon is the implication. The \textbf{Predicted implication} here is from the Claude 3 Sonnet model. } 
\label{tab:example}
\end{table*}


\section{Evaluation}
\subsection{Large Language Models to be Evaluated}
In our evaluation framework, LLMs only interpret the emphasis as conveyed in the input, but do not detect the emphasis. 
We use both open-source and closed commercial LLMs for evaluation. Specifically, we evaluate different versions of Llama~\citep{llama} (Llama 2~\citep{llama2} and Llama 3~\citep{llama3}) and various model sizes (ranging from 7B to 70B parameters). In addition to Llama, we experiment with Mistral 7B~\citep{mistral}. All models used are the instruction-tuned chat versions, not the pre-trained models. For commercial LLMs, we evaluate Claude 3 Sonnet\footnote{https://www.anthropic.com/news/claude-3-family} and ChatGPT 3.5 (\texttt{gpt-3.5-turbo-0125})\footnote{https://openai.com/index/chatgpt/}. For all the LLMs, we provide the following prompt to generate the implied meaning of the emphasized sentence, where \texttt{\{CONTEXT\}} and \texttt{\{CURRENT\}} are placeholders that vary for each sample.\\
\texttt{[Dialogue context]: \{CONTEXT\}}\\
\texttt{[Current turn with emphasis]: \{CURRENT\}} \\
\texttt{The emphasized information is indicated by the quotation mark "". Implication meaning refers to the key intention the speaker wants to specifically highlight beyond the original text, which is not simply paraphrasing the original text. Use a simple and concise sentence to describe the specific highlighted information. Directly output the implication meaning of the current turn sentence.}

\subsection{Evaluation Metrics}
For human evaluation, we recruit human annotators to rate the scoring from 0 to 5, given the rating guideline (the detailed instruction is shown in Appendix Figure \ref{fig:human-eval}). Each sample is rated by three annotators, the average score is the Mean Opinion Score (MOS). 

However, human evaluation incurs high costs and significant time requirements, requiring skilled evaluators who can understand and analyze the subtle nuances of language and context within the dialogues. Therefore, we propose using the state-of-the-art LLM, GPT-4~\citep{gpt4}, for automatic evaluation. The automatic evaluation methods are illustrated in Figure \ref{fig:eval}.
We adopt a similar LLM automatic evaluation method as in previous works~\citep{llm-evaluation, closer}, instructing the LLM to predict ratings based on guidelines. This is denoted as \textbf{auto-gpt4} score. 
Given potential uncertainties in the LLM's comprehension of emphasized sentences, we also use a similarity score compared to \textit{ground truth} annotations, employing both the BERTScore (\textbf{BERT$_{f1}$}) and GPT-4 (\textbf{auto-gpt4-gt}). 
BERTScore~\citep{bertscore} is a metric for evaluating text generation quality by comparing generated texts to reference texts using BERT embeddings. Unlike traditional metrics like BLEU, which rely on exact matches, BERTScore uses contextual embeddings to capture semantic similarities. As for \textbf{auto-gpt4-gt}, 
Appendix \ref{app:scoring} describes more details on the prompt template for GPT-4 evaluation.

\begin{table}[t]
\centering
\begin{tabular}{lcc}
\toprule
\textbf{Auto eval} & \textbf{Pearson's}  & \textbf{Kendall's} \\ \midrule
BERT$_{f1}$                     &   0.313                                    &       0.203             \\
auto-gpt4-gt                  & 0.568                           & 0.268              \\
auto-gpt4                     & \textbf{0.643}                            & \textbf{0.327}             \\ \bottomrule
\end{tabular}
\caption{The Pearson's and Kendall's correlation coefficient between human MOS and of three automatic evaluation methods.}
\label{tab:correlation}
\end{table}

\section{Results}

\subsection{LLMs' Performance on Implication Modeling}
Table \label{tab:main} reports show the scores of different LLMs. Commercial LLMs generally outperform open-source ones in MOS. The top model is Claude 3 Sonnet with a MOS of 3.73, followed by ChatGPT at 3.59. Among open-source LLMs, Llama 3-70B scores 3.51, close to ChatGPT. The Llama 3 series consistently outperforms the Llama 2 series of similar sizes. Mistral 7B ranks between Llama 2-7B and Llama 3-8B, even surpassing Llama 2-70B. None of the LLMs reach the optimal score of 5, with the best at 3.73, indicating room for improvement in understanding emphasized sentences. More error analyses are shown in Appendix \ref{app:error}.

\subsection{Correlation Between Automatic Evaluation Score and Human Score}
We follow ~\citet{closer} to calculate the dataset-level Pearson's correlation and document-level Kendall's correlation. Pearson's measures overall alignment between automatic scores and human ratings, while Kendall's assesses quality differentiation between models for the same input. Table \ref{tab:correlation} shows the correlations of various evaluation methods. Auto-gpt4 scores correlate well with human ratings, with a Pearson's coefficient of 0.643 and a Kendall’s coefficient of 0.327. Both auto-gpt4-gt and BERT$_{f1}$ also show positive correlations, but their coefficients are much lower. This suggests that GPT-4's direct analysis is more effective than using ground truths as references, which can be influenced by human annotators' writing styles and biases. Therefore, the auto-gpt4 score is more reliable for a higher correlation with human ratings.


\subsection{Relative Ranking for Automatic Evaluation Methods}

The ranking order of LLMs by automatic evaluation scores aligns closely with human scores. Spearman's rank correlation between human MOS and three automatic methods shows high coefficients: 0.964 for auto-gpt4 and BERT$_{f1}$, and 0.857 for auto-gpt4-gt. Auto-gpt4 and BERT$_{f1}$ had only two incorrect model rankings. Despite low Pearson's and Kendall's correlations, BERT$_{f1}$'s rankings are similar to human MOS. Overall, auto-gpt4 achieves both the highest score and rank correlation.

\subsection{Qualitative Example}
Table \ref{tab:example} shows examples of the same dialogue context with different emphasized words or phrases. For the emphasized word \textit{single}, the Claude 3 Sonnet model incorrectly interprets it as referring to a cart for individual use, whereas it actually refers to the number of carts. For the emphasized phrase \textit{right now}, the model correctly understands it implies immediate use. This demonstrates that while commercial LLMs can infer different meanings based on emphasis, their accuracy depends on the complexity and depth of the context.


\section{Conclusion}
This work explores the ability of LLMs to interpret emphasized text in dialogues. We introduced the novel Emphasized-Talk dataset with the same dialogue context and current turn sentence with different emphasized words or phrases. Furthermore, we study the automatic evaluation methods using GPT-4, showing a high correlation between human rating and auto-gpt4. Our analysis shows that while commercial LLMs outperform open-sourced models, there is still room for improvement. We encourage future LLM research to evaluate the model on the proposed benchmark and automatic evaluation method to enhance the LLM's dialogue ability.

\section*{Limitations}
The primary limitations of this paper are as follows: 
\begin{enumerate}
    \item This study exclusively utilizes GPT-4 for automatic evaluation, relying on previous works~\citep{closer, geval} that have demonstrated its effectiveness for scoring. However, the performance of other LLMs for scoring is not explored in this paper.
    \item While speech emphasis is a common feature of human communication, this work simplifies emphasis by using additional quotation marks in text. Future research should investigate the use of direct speech input for spoken dialogue to better capture natural communication.
    \item The study employs quotation marks to indicate emphasis within text sentences and explicitly instructs that the emphasized information is enclosed by these marks. However, other methods of emphasis, such as bold text or capitalization, are not explored in this paper.
\end{enumerate}

\section*{Acknowledgement}
The authors want to thank Cheng-Han Chiang for his insightful discussion. Guan-Ting Lin is supported by the NTU GICE and the NTU Ph.D. scholarship. 



\bibliography{custom}

\appendix
\section*{Appendix}

\begin{table*}[t]
\centering
\begin{tabularx}{\textwidth}{>{\raggedright\arraybackslash}X}
\toprule
\textbf{Failure cases} \\
\midrule
0: can anyone tell me where I can buy the ingredients to make Chinese food? \\
1: There is a supermarket in Kensington High Street. I'll come and give you a hand. \\
0: That is "good" of you. [\texttt{\textbf{current turn}}] \\
\textbf{Claude 3 Sonnet's Prediction:} The speaker is expressing sarcasm or lack of sincerity in response to the offer of help. \\
\textbf{Llama 2-7B's Prediction:} The speaker is offering to help the listener find the supermarket and purchase the ingredients needed for Chinese cooking. \\
\textbf{Mistral-7B's Prediction:} The speaker is offering to help the person find ingredients for Chinese food. \\
\bottomrule
\end{tabularx}
\caption{Failure cases in which the model prediction receives a low rating from the human.}
\label{tab:fail}
\end{table*}

\section{Analysis on Unsatisfied Model Prediction}
\label{app:error}
We analyze the cases resulting in unsatisfactory ratings (average MOS score below 2). The examples are shown in Table \ref{tab:fail}. We observe that Claude 3 Sonnet overinterprets the emphasized information in an extremely contrasting way, but this is not relevant to the dialogue context. In contrast, the Llama 2-7B and Mistral-7B models fail to focus on the current turn and the emphasized word, merely summarizing the entire dialogue. These results indicate that small open-sourced LLMs struggle to understand emphasized text and its implications, and sometimes even fail to follow instructions. On the other hand, commercial LLMs can follow instructions but may overinterpret the emphasized text. The ideal interpretation should be reasonable and relevant to the dialogue context.

\section{Prompt for Checking Agreement}
\label{appendix:agreement_check}
The prompt for checking agreement is: \texttt{For each sample, given a set of text sentences, check whether all the following sentences are semantically close to each other or not. You should also consider differences in subtle and nuanced meanings. You should provide an explanation, and then output yes if all sentences are semantically close to each other; otherwise, output no.}


\section{Inter-annotator Agreement for Human Evaluation Score}
To measure the inter-annotator agreement, we use Krippendorff's alpha score\footnote{https://github.com/pln-fing-udelar/fast-krippendorff}. From the rating of all the models, Krippendorff's alpha score is 0.255, which means moderate agreement among annotators. The result is reasonable since there is certain freedom for the implication meaning, depending on the background and personality of each annotator.

\section{Details of Automatic Evaluation}
\label{app:scoring}
\begin{figure*}{}
    \centering
    \begin{mdframed}
\includegraphics[width=1\linewidth]{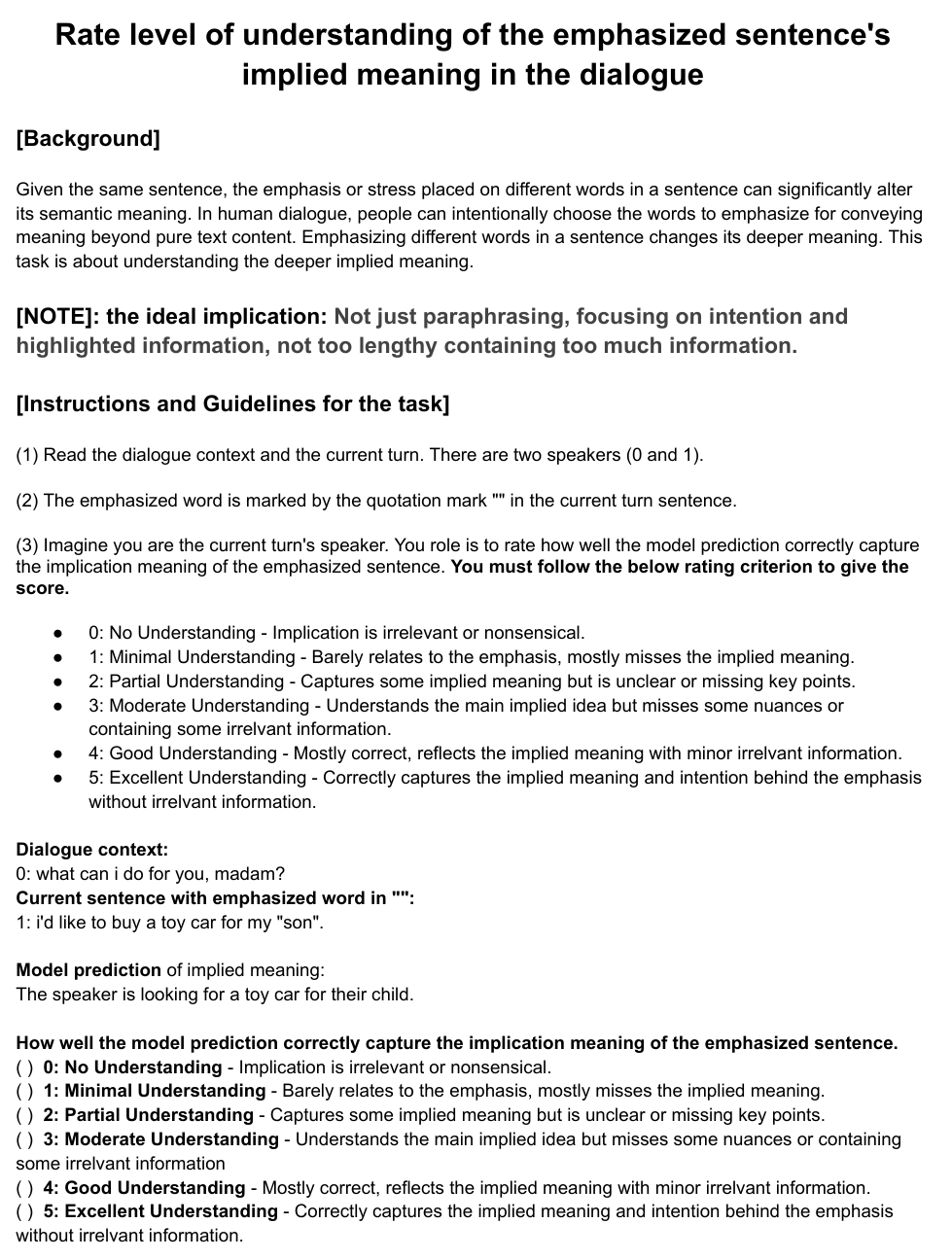}
    \end{mdframed}
    \caption{The template for human evaluation, including instructions and grading policy.}
    \label{fig:human-eval}
\end{figure*}

Following \citet{closer}, we first ask the GPT-4 to analyze input samples and then predict scores. We request GPT-4 to output 3 outputs and average the score to reduce the score variation due to randomness. For the auto-gpt4 score, we use a similar prompt similar as the instruction in Figure \ref{fig:human-eval}. For the auto-gpt4-gt score, GPT-4 rates the similarity between ground truths and model predictions, using task instructions, grading definitions, dialogue context, and the emphasized current turn sentence. The prompt is as follows: \\\\
\texttt{The task is modeling the implication meaning of the emphasized sentence. We are checking how semantically close and subtle meaning differences between the ground truth sentences and model prediction. For the subtle and nuanced meaning, focusing on the intention and highlighted information of the speaker. 
You must follow the following steps to provide the score: 
First, analyze and explain the sentences with the above definition. 
Second, output just the number of scores from the range of integers from 0 to 5: \\
0: No semantic similarity; the model prediction completely diverges from the ground truth in meaning and nuance.\\
1: Very low semantic similarity; only a few elements match, with significant differences in meaning and nuance.\\
2: Low semantic similarity; some parts match, but there are notable differences in meaning and nuance.\\
3: Moderate semantic similarity; many parts match, but some differences in meaning and nuance are present.\\
4: High semantic similarity; most parts match, with minor differences in meaning and nuance.\\
5: Perfect or near-perfect semantic similarity; the model prediction closely mirrors the ground truth in both meaning and nuance.\\
The response must be in valid JSON format as below, which can be correctly parsed by json.loads() in python: 
{"analysis": explanation, "score": number}
}

For the BERT score, we average the scores if multiple ground truths are available\footnote{https://huggingface.co/spaces/evaluate-metric/bertscore}.

\section{Dataset License}
We release the Emphasized-Talk dataset under the MIT license.

\section{Details of Human Annotation Process}
We assign three annotators for each assignment. All are based in the United States with HIT approval rates higher than 98\%, given that the corpus is in American English. Each test contains 20 samples for evaluation. We pay the annotators 2.5 USD for each test. On average, based on the time for annotating and reading the content, it takes 10 minutes on one test,
so the hourly wage is around 15 USD.

\begin{figure*}{}
    \centering
    \begin{mdframed}
\includegraphics[width=1\linewidth]{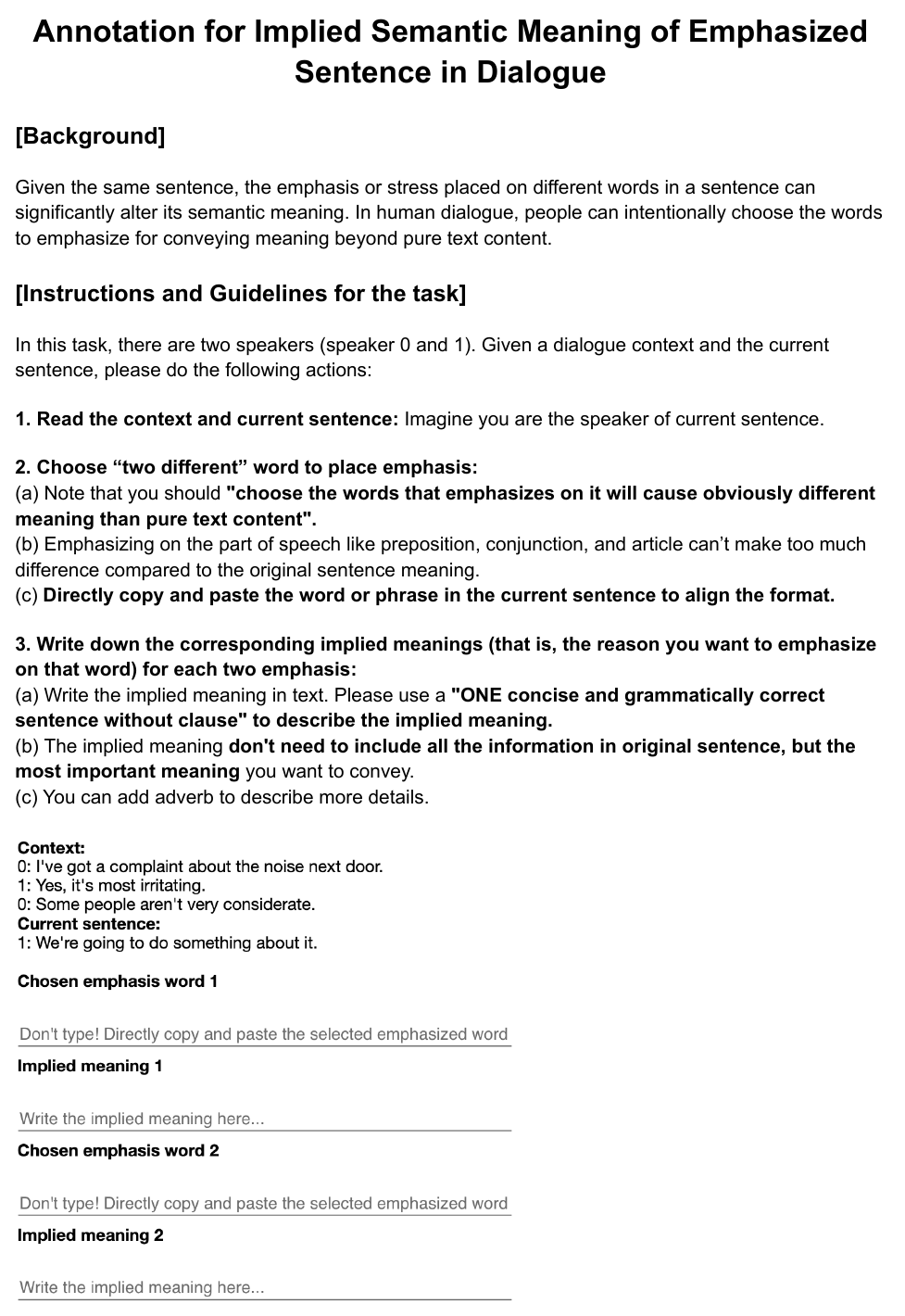}
    \end{mdframed}
    \caption{The template for selecting emphasized words and documenting their implied meanings.}
    \label{fig:annotation}
\end{figure*}

\begin{table*}[t]
\centering
\adjustbox{width=1\textwidth}{
\begin{tabular}{ll}
\toprule
\multicolumn{2}{l}{\textit{\textbf{Example 1}}}                                                                     \\ \hline
Input  & \begin{tabular}[c]{@{}l@{}}1. The speaker is expressing genuine happiness and appreciation for the compliment about their country. \\ 2. I am genuinely happy to hear that you love japan. \\ 3. I express my happiness upon hearing your admiration for japan.\end{tabular} \\ \hline
Output & \begin{tabular}[c]{@{}l@{}}Output: Yes\\ Explanation: All three sentences convey the speaker's genuine happiness in response to a compliment \\ about Japan. They all express positive feelings related to the admiration of Japan.\end{tabular}                             \\ \hline
\multicolumn{2}{l}{\textit{\textbf{Example 2}}}                                                                                                                                                                                                                                                         \\ \hline
Input  & \begin{tabular}[c]{@{}l@{}}1. You should have expressed your deep feelings for her to me. \\  2. The relationship is serious and significant. \\  3. You should have informed me about your romantic feelings for her.\end{tabular}                                          \\ \hline
Output & \begin{tabular}[c]{@{}l@{}}Output: No \\ Explanation: The second sentence discusses the seriousness of the relationship in general, which \\ differs from the specific recommendation of expressing feelings in the first and third sentences.\end{tabular}                  \\ \bottomrule
\end{tabular}}
\caption{The examples of filtering by agreement process.}
\label{tab:filtering_agreement}
\end{table*}

\begin{table*}[]
\centering
\adjustbox{width=1\textwidth}{
\begin{tabular}{ll}
\toprule
\multicolumn{2}{l}{\textit{Example of auto-gpt4-gt score}}                                                                                                                                                                                                                                                             \\ \hline
Input  & \begin{tabular}[c]{@{}l@{}}\textbf{Dialogue context}: What sort of music do you like? \\ \textbf{Current turn}: I love just about “anything” by prince.\\ \textbf{Model prediction}: The speaker has a strong preference for and enjoyment of Prince's music.\\ \textbf{Ground truth}: The speaker's affection for prince's music encompasses a wide range, indicating \\ a broad and inclusive taste.\end{tabular}                                                                                                                                                                                                       \\ \hline
Output & \begin{tabular}[c]{@{}l@{}}\textbf{Analysis}: The ground truth emphasizes the breadth of the speaker's affection towards Prince's music, \\ highlighting an inclusive and wide-ranging taste. The model prediction focuses more narrowly, noting \\ the speaker's strong preference and enjoyment, but doesn't capture the aspect of 'broad and inclusive taste'. \\ Therefore, while both sentences refer to the speaker's positive feelings towards Prince's music, the model\\ prediction lacks the subtlety about the speaker's taste being wide and inclusive.\\ \textbf{Score}: 3\end{tabular} \\ \bottomrule
\end{tabular}}
\caption{Example input and output for the \texttt{auto-gpt4-gt} score. }
\label{tab:ex_auto_gpt4_gt}
\end{table*}

\end{document}